\pdfoutput=1
\documentclass[11pt]{article}

\usepackage[]{naacl2022}

\usepackage{times}
\usepackage{latexsym}

\usepackage{amsmath}
\usepackage{mathtools}
\usepackage{tikz}
\usepackage{mathdots}
\usepackage{yhmath}
\usepackage{adjustbox}
\usepackage{times}
\usepackage{latexsym}
\usepackage{multirow}
\usepackage{graphicx}
\usepackage{booktabs}
\usepackage{nccmath}
\usepackage{bbm}

\usepackage{physics}
\usepackage{amsmath}
\usepackage{tikz}
\usepackage{mathdots}
\usepackage{yhmath}
\usepackage{cancel}
\usepackage{color}
\usepackage{array}
\usepackage{multirow}
\usepackage{amssymb}
\usepackage{tabularx}
\usepackage{extarrows}
\usepackage{booktabs}
\usepackage{algorithm}
\usepackage{algpseudocode}
\usepackage{graphicx}
\usetikzlibrary{fadings}
\usetikzlibrary{patterns}
\usetikzlibrary{shadows.blur}
\usetikzlibrary{shapes}

\makeatletter
\newcommand{\multiline}[1]{%
  \begin{tabularx}{\dimexpr\linewidth-\ALG@thistlm}[t]{@{}X@{}}
    #1
  \end{tabularx}
}
\makeatother

\algblock{Input}{EndInput}
\algnotext{EndInput}
\algblock{Output}{EndOutput}
\algnotext{EndOutput}

\usepackage[T1]{fontenc}
\usepackage[utf8]{inputenc}
\usepackage{microtype}

\title{A Data Cartography based MixUp for Pre-trained Language Models}

\author{Seo Yeon Park and Cornelia Caragea\\
Computer Science  \\
  University of Illinois Chicago \\
  {\tt spark313@uic.edu} \mbox{     } \mbox{     } \mbox{     }
    {\tt cornelia@uic.edu}
}

\begin{document}
\maketitle
\begin{abstract}
MixUp is a data augmentation strategy where additional samples are generated during training by combining \textit{random pairs} of training samples and their labels. However, selecting random pairs is not potentially an optimal choice. In this work, we propose TDMixUp, a novel MixUp strategy that leverages \textbf{T}raining \textbf{D}ynamics and allows more informative samples to be combined for generating new data samples. Our proposed TDMixUp first measures confidence, variability, \cite{swayamdipta2020dataset}, and Area Under the Margin (AUM) \cite{pleiss2020identifying} to identify the characteristics of training samples (e.g., as easy-to-learn or ambiguous samples), and then interpolates these characterized samples. We empirically validate that our method not only achieves competitive performance using a smaller subset of the training data compared with strong baselines, but also yields lower expected calibration error on the pre-trained language model, BERT, on both in-domain and out-of-domain settings in a wide range of NLP tasks. We publicly release our code.\footnote{\href{https://github.com/seoyeon-p/TDMixUp}{https://github.com/seoyeon-p/TDMixUp}}
\end{abstract}

\section{Introduction}
MixUp \cite{zhang2017mixup} is a simple data augmentation strategy in which additional samples are generated during training by combining \textit{random pairs} of training samples and their labels. 
While simple to implement, MixUp has been shown to improve both predictive performance and model calibration (i.e., avoiding over-confident predictions) \cite{guo2017calibration} due to its regularization effect through data augmentation \cite{singh2019mixup}. 
However, selecting \textit{random pairs} in MixUp might not necessarily be optimal.

Despite this, MixUp has been explored for NLP tasks with substantial success using hidden state representations \cite{verma2019manifold}. 
For instance, \citet{sun-etal-2020-mixup} explored MixUp, which uses the hidden representation of BERT \cite{devlin2019bert} to synthesize additional samples from randomly selected pairs.
\citet{yin-etal-2021-batchmixup} proposed MixUp, which uses the hidden representation of RoBERTa \cite{liu2019roberta} to interpolate all samples in the same mini-batch to better cover the feature space.
To date, only a few prior works have focused on selecting informative samples for MixUp. For example, \citet{chen-etal-2020-mixtext} proposed semi-supervised learning, which interpolates labeled and unlabeled data based on entropy. \citet{kong-etal-2020-calibrated} explored BERT calibration with MixUp, which generates new samples by exploiting the distance between samples in the feature space. 

Recently, \citet{swayamdipta2020dataset} introduced data maps, which allow evaluating data quality by using training dynamics (i.e., the behavior of a model as training progresses). 
Specifically, they consider the mean and standard deviation of the gold label probabilities, predicted for each sample across training epochs (i.e., confidence and variability), and characterize data into three different categories: (1) samples that the model predicts correctly and consistently (i.e., \textbf{easy-to-learn}); (2) samples where true class probabilities vary frequently during training (i.e., \textbf{ambiguous}); and (3) samples that are potentially mis-labeled or erroneous (i.e., \textbf{hard-to-learn}). 
The author revealed that the easy-to-learn samples are useful for model optimization (parameter estimation) and without such samples the training could potentially fail to converge, while the ambiguous samples are those on which the model struggles the most and push the model to become more robust, hence, these ambiguous samples are the most beneficial for learning since they are the most challenging for the model. 

Inspired by these observations, we propose a novel MixUp strategy which we call {TDMixUp} that monitors training dynamics and interpolates easy-to-learn samples with ambiguous samples in the feature space. 
That is, we pair one sample from the easy-to-learn set with another sample from the ambiguous set to allow more informative samples to be combined for MixUp. 
Accordingly, we generate new samples that share the characteristics of both easy-to-learn and ambiguous data samples and are hence more beneficial for learning.
However, the easy-to-learn and the ambiguous sets can contain mis-labeled samples that can degrade the model performance.
Consequently, we measure another training dynamic, Area Under the Margin (AUM) \cite{pleiss2020identifying}, to filter out possibly mis-labeled samples in each set. 
We validate our proposed method on a wide range of natural language understanding tasks including textual entailment, paraphrase detection, and commonsense reasoning tasks. We achieve competitive accuracy and low expected calibration error \cite{guo2017calibration} on both in-domain and out-of-domain settings for the pre-trained language model BERT \cite{devlin2019bert}, without using the full training data.

\section{Proposed Approach: TDMixUp}
We introduce our proposed TDMixUp, which generates additional samples based on the characteristics of the data samples. We first reveal the characteristics of each data sample by using training dynamics, i.e., confidence, variability, and Area Under the Margin (AUM). We then describe our MixUp operation that combines training samples based on the above data characteristics that are measured during training.

\subsection{Data Samples Characterization}
\label{sec:char}
We first introduce \textbf{confidence} and \textbf{variability}, that are used to evaluate the characteristics of each individual sample \cite{swayamdipta2020dataset}. 
These statistics are calculated for each sample $(x_i,y_i)$ over $E$ training epochs. 
\paragraph{Confidence} We define confidence as the mean model probability of the true label $y_i$ across epochs:
\begin{equation}
    \hat{\mu}_i = \frac{1}{E}\sum_{e=1}^{E}p_{\theta^{(e)}}(y_i|{x}_i)
\end{equation}
where $p_{\theta^{(e)}}$ denotes the model's probability with parameter $\theta^{(e)}$ at the end of $e^{th}$ epoch. 
\paragraph{Variability} We define variability as the standard deviation of $p_{\theta^{(e)}}$ across epochs $E$:
\begin{equation}
    \hat{\sigma}_i = \sqrt{\frac{\sum_{e=1}^{E}(p_{\theta^{(e)}}(y_i|{x}_i)-\hat{\mu}_i)^2}{E}}
\end{equation}
Given these statistics per sample, we identify the top 33\% easy-to-learn samples, i.e., those samples that the model predicts correctly and consistently across epochs (high-confidence, low-variability), and the top 33\% ambiguous samples, i.e., those samples whose true class probabilities have a high variance during training (high-variability).

\paragraph{Area Under the Margin (AUM)}
As another measure of data quality, we monitor training dynamics using the Area Under the Margin (AUM) \cite{pleiss2020identifying}.
AUM measures how different a true label for a sample is compared to a model's belief at each epoch and is calculated as the average difference between the logit values for a sample's assigned class (gold label) and its highest non-assigned class across training epochs. 
Formally, given a sample $({x}_i,y_i)$, we compute $AUM({x}_i,y_i)$ as the area under the margin averaged across all training epochs $E$. 
Specifically, at some epoch $e\in E$, the margin is defined as:
\begin{equation}
    M^{e}(x_i, y_i) =  z_{y_i} - max_{y_i!=k}(z_{k})
\end{equation}
where $M^{e}(x_i, y_i)$ is the margin of sample $x_i$ with true label $y_i$, $z_{y_i}$ is the logit corresponding to the true label $y_i$, and $max_{y_i!=k}(z_{k})$ is the largest {\em other} logit corresponding to label $k$ not equal to $y_i$. 
The AUM of $(x_i,y_i)$ across all epochs is:
\begin{equation}
    AUM(x_i, y_i) =  \frac{1}{E} \sum_{e=1}^{E}M^{e}(x_i, y_i)
\end{equation}
\noindent 
Intuitively, while both AUM and confidence measure training dynamics, confidence simply measures the probability output of the gold label and how much it fluctuates over the training epochs.
In contrast, AUM measures the probability output of the gold label with respect to the model's belief in what the label for a sample should be according to its generalization capability (derived by observing other similar samples during training). More precisely, AUM considers each logit value and measures how much the gold label assigned logit value differs from {\em the other largest} logit value, which allows identifying mis-labeled samples. 

To identify possibly mis-labeled data in each set (i.e., the set of easy-to-learn and the set of ambiguous samples that are categorized by confidence and variability as described above), we first fine-tune a model on each set, respectively, with inserting fake data (i.e., threshold samples).
Data with similar or worse AUMs than threshold samples can be assumed to be mis-labeled \cite{pleiss2020identifying}. We construct threshold samples by taking a subset of the training data and re-assigning their labels randomly, including a class that does not really exist. 
Specifically, given $N$ training samples that belong to $c$ classes, we randomly select $N/(c+1)$ samples per class and re-assign their labels to classes that are different from the original class. 
We then train a model on training samples including threshold samples and measure the AUMs of all training data. We identify possible mis-labeled data by computing a threshold value (i.e., the $k$th percentile threshold sample AUMs where $k$ is a hyper-parameter chosen on the validation set). 
At last, we filter out samples that have lower AUM than the threshold value.

\begin{table*}[hbt!]
\small
\centering
\begin{tabular}{lcccccc}
\toprule
                              & \multicolumn{2}{c}{SNLI}                               & \multicolumn{2}{c}{QQP}                                & \multicolumn{2}{c}{SWAG}                               \\ 
                              \cmidrule(lr){2-3} 
                              \cmidrule(lr){4-5}
                              \cmidrule(lr){6-7}
                              & \multicolumn{1}{c}{Acc} & \multicolumn{1}{c}{ECE} & \multicolumn{1}{c}{Acc} & \multicolumn{1}{c}{ECE} & \multicolumn{1}{c}{Acc} & \multicolumn{1}{c}{ECE} \\\midrule
100\% train                          & $\textbf{90.04}_{0.3}$                        & $2.54_{0.8}$                    & $\textbf{90.27}_{0.3}$                       & $2.71_{0.5}$                    & $\textbf{79.40}_{0.4}$                        & $2.49_{1.8}$                    \\ \midrule
33\% train, Easy-to-learn            & $82.78_{0.6}$                        & $16.22_{0.7}$                   &$63.16_{0.1}$	&$36.88_{0.1}$                  & $75.39_{0.2}$                        & $17.51_{0.1}$                   \\
24\% train, Easy-to-learn with AUM &$83.03_{0.9}$	&$15.05_{0.9}$	 & $66.43_{0.6}$            & $33.93_{0.8}$ 			&$75.56_{0.1}$	&$15.81_{0.7}$ \\
33\% train, Ambiguous                & $89.71_{0.5}$                        & $\textbf{0.74}_{0.1}$                    & $87.51_{0.5}$                        & ${1.71}_{0.4}$                    & $75.91_{0.6}$                        & $\textbf{1.84}_{0.7}$                    \\ 
24\% train, Ambiguous with AUM                             & $87.88_{0.7}$                        & $7.09_{0.8}$                   & $88.63_{0.5}$                        & ${6.36}_{0.6}$                   &$71.74_{0.4}$                         &$7.55_{1.1}$                   \\ 
66\% train, Easy-to-learn \& Ambiguous     & $89.65_{0.2}$                        & $2.64_{0.5}$                    & $90.23_{0.7}$                        & $\textbf{1.35}_{0.4}$                    & $78.78_{0.5}$                             & $2.51_{0.8}$                        \\

\midrule \midrule
                                           & \multicolumn{2}{c}{MNLI}                               & \multicolumn{2}{c}{TwitterPPDB}                        & \multicolumn{2}{c}{HellaSWAG}                          \\ 
                                           
                              \cmidrule(lr){2-3} 
                              \cmidrule(lr){4-5}
                              \cmidrule(lr){6-7}
                                           & \multicolumn{1}{c}{Acc} & \multicolumn{1}{c}{ECE} & \multicolumn{1}{c}{Acc} & \multicolumn{1}{c}{ECE} & \multicolumn{1}{c}{Acc} & \multicolumn{1}{c}{ECE} \\\midrule
100\% train                                       &${73.52}_{0.3}$                        & ${7.09}_{2.1}$                    & $\textbf{87.63}_{0.4}$                        & $8.51_{0.6}$                    & $\textbf{34.48}_{0.2}$                        & $12.62_{2.8}$                   \\ \midrule
33\% train, Easy-to-learn                          & $61.41_{0.8}$                        & $36.68_{1.9}$ &$81.07_{0.8}$	&$18.92_{0.7}$                                     & $33.59_{1.1}$                        & $29.38_{2.1}$                   \\
24\% train, Easy-to-learn with AUM &$62.97_{1.5}$	&$32.48_{2.9}$		& $82.16_{0.7}$                        & $17.46_{1.0}$	&	$33.67_{1.4}$	&$16.89_{2.6}$ \\
33\% train, Ambiguous                             & $72.52_{1.2}$                        & $10.73_{1.0}$                   & $86.62_{0.6}$                        & $\textbf{6.01}_{1.1}$                   & $34.29_{0.9}$                        & ${8.40}_{1.3}$                   \\ 
24\% train, Ambiguous with AUM      &$70.87_{0.9}$ & $17.23_{1.6}$ & $86.59_{0.8}$ &$7.31_{0.8}$    &$33.81_{1.0}$  &$\textbf{3.76}_{2.3}$                                       \\ 
66\% train, Easy-to-learn \& Ambiguous            & $\textbf{73.89}_{0.6}$                        & $\textbf{3.46}_{1.9}$                    & $87.29_{0.3}$                        & $8.04_{0.7}$                    & $34.43_{0.2}$                        & $9.68_{1.1}$                    \\ 
\midrule
\end{tabular}
\caption{\small{The comparison of accuracy and expected calibration error (ECE) in percentage on in-domain (top) and out-of-domain (bottom) for BERT. We compare the results of fine-tuning on subsets of train samples (i.e., easy-to-learn and ambiguous samples) and fine-tuning on the entire 100\% of training samples. Lower ECE implies  better calibrated models. 
We report the mean accuracy across five training runs with the standard deviation shown in subscript (e.g., $90.04_{0.3}$ indicates $90.04 \pm 0.3$).}}
\label{tb:1}
\end{table*}

\subsection{MixUp}
\label{sec:mixup}
MixUp training generates vicinity training samples according to the rule introduced in \citet{zhang2017mixup}:  
\begin{equation}
\begin{aligned}
    \tilde{x} = \lambda x_{i} + (1-\lambda)x_{j} \\
    \tilde{y} = \lambda y_{i} + (1-\lambda)y_{j}
\end{aligned}
\label{eq:1}
\end{equation}
where $x_{i}$ and $x_{j}$ are two randomly sampled input points, $y_{i}$ and $y_{j}$ are their associated one-hot encoded labels, and $\lambda$ is a mixing ratio sampled from a Beta($\alpha$, $\alpha$) distribution with a hyper-parameter $\alpha$.
In standard MixUp, training data is augmented by linearly interpolating \textit{random} training samples in the input space. 
In contrast, our proposed TDMixUp interpolates one easy-to-learn sample with one ambiguous sample after applying AUM to filter potential erroneous samples that harm performance.
Our current implementation uses easy-to-learn and ambiguous data loaders respectively and then applies MixUp to a randomly sampled mini-batch of each loader. We train a model on the generated TDMixUp samples in addition to the easy-to-learn and ambiguous samples using the cross entropy-loss.

\section{Experiments and Results}
\subsection{Tasks and Datasets}
We evaluate our TDMixUp on three natural language understanding tasks.
We describe our in-domain and out-of-domain sets as follows. 

\paragraph{Natural Language Inference (NLI)}
Stanford Natural Language Inference (SNLI) is a task to predict if the relation between a hypothesis and a premise is \textit{entailment, contradiction,} or \textit{neutral} \cite{bowman-etal-2015-large}.
Multi-Genre Natural Language Inference (MNLI) captures NLI with diverse domains \cite{william2018mnli}.

\paragraph{Paraphrase Detection} 
Quora Question Pairs (QQP) is a paraphrase detection task to test if two questions are semantically equivalent \cite{lyer2017qqp}.
TwitterPPDB (TPPDB) is a dataset built to determine whether sentence pairs from Twitter convey similar semantics when they share URLs \cite{lan-etal-2017-continuously}.

\paragraph{Commonsense Reasoning} 
Situations With Adversarial Generations (SWAG) is a commonsense reasoning task to choose the most plausible continuation of a sentence among four candidates \cite{zellers2018swag}. 
HellaSWAG is a dataset built using adversarial filtering to generate challenging out-of-domain samples.

\subsection{Experimental Setup}
We use BERT \cite{devlin2019bert} based classification model and pass the resulting \texttt{[CLS]} representation through a fully connected layer and softmax to predict the label distribution. We follow the published train/validation/test datasets splits as described in \citet{desai2020calibration}. 
To identify mis-labeled samples in the top 33\% easy-to-learn samples, we set threshold values $k$ as: the 80th/80th/50th percentile threshold sample AUMs on SNLI/QQP/SWAG, respectively.
More training details and hyper-parameter settings can be found in the Appendix.
We evaluate the capability of our TDMixUp strategy to improve both predictive performance and model calibration due to its regularization effect through data augmentation.  
Hence, we use two metrics: (1) accuracy, and (2) expected calibration error (ECE) \cite{guo2017calibration,desai2020calibration}. We report results averaged across 5 fine-tuning runs with random restarts for all experiments.

\subsection{Baseline Methods}
\paragraph{BERT} \cite{devlin2019bert} is the pre-trained base BERT model fine-tuned on each downstream task.
\paragraph{Back Translation Data Augmentation} \cite{edunov2018understanding} generates augmented samples by using pre-trained translation models\footnote{We use FairSeq and set the random sampling temperature as 0.9.} which can generate diverse paraphrases while preserving the semantics of the original sentences. In experiments, we translate original sentences from English to German and then translate them back to English to obtain the paraphrases. 
\paragraph{MixUp} \cite{zhang2017mixup} generates augmented samples by interpolating {random} training samples in the input space (obtained from the first layer of the BERT pre-trained language model).
\paragraph{Manifold MixUp (M-MixUp)} \cite{verma2019manifold} generates additional samples by interpolating random training samples in the feature space (obtained from the task-specific layer on top of the BERT pre-trained language model).
\paragraph{MixUp for Calibration} \cite{kong-etal-2020-calibrated} generates augmented samples by utilizing the cosine distance between samples in the feature space. 

\vspace{3mm}
\noindent Note that the above baselines use 100\% training data while our proposed method focuses on particular subsets of the training data.

\begin{table*}[hbt!]
%\small
\resizebox{\textwidth}{!}{
\centering
\begin{tabular}{lcccccc}
\toprule
                              & \multicolumn{2}{c}{SNLI}                               & \multicolumn{2}{c}{QQP}                                & \multicolumn{2}{c}{SWAG}                               \\ 
                                           
                              \cmidrule(lr){2-3} 
                              \cmidrule(lr){4-5}
                              \cmidrule(lr){6-7}
                              & \multicolumn{1}{c}{Acc} & \multicolumn{1}{c}{ECE} & \multicolumn{1}{c}{Acc} & \multicolumn{1}{c}{ECE} & \multicolumn{1}{c}{Acc} & \multicolumn{1}{c}{ECE} \\
                              \midrule
100\% train                          & $90.04_{0.3}$                        & $2.54_{0.8}$                    & $90.27_{0.3}$                        & $2.71_{0.5}$                    & $79.40_{0.4}$                        & $2.49_{1.8}$                    \\ 
100\% train, MixUp \cite{zhang2017mixup} &$88.82_{0.2}$ &$7.73_{1.1}$ &$89.12_{0.5}$ &$9.04_{0.8}$ &$74.98_{2.3}$ & $7.08_{1.0}$ \\ 
100\% train, M-MixUp \cite{verma2019manifold} &$89.45_{0.9}$ &$1.51_{0.8}$ &$89.93_{0.6}$ & $3.02_{1.0}$  &$78.26_{0.4}$ & $4.12_{0.6}$ \\ 
100\% train, MixUp for Calibration \cite{kong-etal-2020-calibrated} &$89.25_{0.5}$ &$2.16_{0.5}$ &$90.24_{0.3}$ & $5.22_{0.6}$  &$79.44_{0.6}$ & $\textbf{1.10}_{0.4}$ \\ 
100\% train, Back Translation Data Augmentation \cite{edunov2018understanding} &$89.22_{0.5}$ &$1.98_{0.6}$ &$89.18_{0.6}$ &$5.01_{0.3}$ &$76.22_{0.9}$ &$1.24_{0.2}$ \\ 
\midrule
66\% train, TDMixUp, Easy-to-learn + Ambiguous         & $89.73_{0.1}$                        & $2.39_{0.8}$                    & $89.77_{0.2}$                        & $1.89_{0.4}$                    & $78.38_{0.3}$                        & $4.21_{0.3}$                    \\ 
{57\% train, TDMixUp, Easy-to-lean with AUM + Ambiguous (Ours)} & $\textbf{90.31}_{0.2}$               & $\textbf{1.22}_{0.4}$           & $\textbf{90.42}_{0.2}$               & $\textbf{1.53}_{0.9}$           & $\textbf{79.59}_{0.3}$               & {$2.16_{0.4}$} \\ \midrule \midrule
                                           & \multicolumn{2}{c}{MNLI}                               & \multicolumn{2}{c}{TwitterPPDB}                        & \multicolumn{2}{c}{HellaSWAG}                          \\ 
                                           
                              \cmidrule(lr){2-3} 
                              \cmidrule(lr){4-5}
                              \cmidrule(lr){6-7}
                                                                                     & \multicolumn{1}{c}{Acc} & \multicolumn{1}{c}{ECE} & \multicolumn{1}{c}{Acc} & \multicolumn{1}{c}{ECE} & \multicolumn{1}{c}{Acc} & \multicolumn{1}{c}{ECE} \\\midrule
100\% train                                        & $73.52_{0.3}$                        & $7.09_{2.1}$                    & $87.63_{0.4}$                        & $8.51_{0.6}$                    & $34.48_{0.2}$                        & $12.62_{2.8}$                   \\ 
100\% train, MixUp \cite{zhang2017mixup} &$69.19_{0.8}$ &$19.51_{2.1}$ &$87.45_{0.3}$ &$11.70_{1.6}$  &$33.22_{0.4}$ & $10.93_{2.0}$ \\ 
100\% train, M-MixUp \cite{verma2019manifold} &$73.22_{0.6}$ &$8.06_{1.2}$ &$87.58_{0.7}$ & $7.68_{1.3}$  &$34.86_{0.9}$ &$13.56_{1.6}$ \\ 
100\% train, MixUp for Calibration \cite{kong-etal-2020-calibrated} &$64.90_{0.5}$ &$17.75_{1.8}$ &$74.51_{1.1}$ & $11.83_{1.0}$  &$32.51_{0.8}$ & $31.61_{2.3}$ \\ 
100\% train, Back Translation Data Augmentation \cite{edunov2018understanding} &$73.15_{0.7}$ &$8.46_{1.3}$ &$86.82_{0.7}$ &$8.83_{0.6}$ &$34.97_{0.4}$ &$22.68_{3.3}$ \\ 
\midrule
66\% train, TDMixUp, Easy-to-learn  + Ambiguous     & $72.83_{1.1}$                        & $5.84_{1.9}$                    & $87.63_{0.2}$                        & $6.48_{0.7}$                    & $34.11_{0.1}$                        &$10.54_{1.6}$                   \\ 
{57\% train, TDMixUp, Easy-to-learn with AUM + Ambiguous (Ours)} & $\textbf{74.28}_{0.6}$               & $\textbf{2.91}_{1.4}$           & $\textbf{87.89}_{0.3}$               & $\textbf{6.08}_{0.4}$           & $\textbf{35.21}_{0.6}$               & $\textbf{9.45}_{1.3}$           \\ \bottomrule
\end{tabular}
}
\caption{\small{Accuracy (in percentage) and expected calibration error (ECE) on in-domain (top) and out-of-domain (bottom) for BERT when comparing our proposed TDMixUp with baseline methods. Bold text shows the best performance and calibration. For \textit{100\% train} results, we use reported results by \citet{desai2020calibration}. We report the mean accuracy across five training runs with the standard deviation shown in subscript (e.g., $90.04_{0.3}$ indicates $90.04 \pm 0.3$).}}
\label{tb:2}
\end{table*}

\subsection{Results}
\paragraph{Fine-tuning on Subsets of Training Data} To explore the effect of different subsets of the data that are characterized using training dynamics, we compare the result of BERT fine-tuned on 100\% training data with the results of BERT when fine-tuned on these subsets, and show the comparison in Table \ref{tb:1}. 
Note that, for each task, we train the model on in-domain training set, and evaluate on in-domain and out-of-domain test sets. We make the following observations: 
First, we observe that accuracy and ECE improve when we filter out possibly mis-labeled samples in the top 33\% easy-to-learn samples by using AUM in all cases (for both in-domain and out-of-domain test sets). Specifically, using \textit{24\% train, Easy-to-learn with AUM}\footnote{Although we write 24\% in Table \ref{tb:1} for simplicity, the percentage of training samples on SNLI/QQP/SWAG after AUM filtering are 20\%/24\%/25\%, respectively} returns better accuracy and lower ECE than \textit{33\% train, easy-to-learn}, showing that there are some potentially erroneous samples that harm the performance in the top 33\% easy-to-learn samples. 
We manually investigate filtered samples on the top 33\% easy-to-learn samples and observe that the top 33\% easy-to-learn samples indeed include mis-labeled samples. For example, in SNLI, we observe that the relation between the following pairs of sentences in the top 33\% easy-to-learn samples is {\em contradiction} when it should be {\em neutral}: \textit{<Two opposing wrestlers competing to pin one another.; `Two women are shopping in a boutique.'>} and \textit{<`A person dressed in a colorful costume is holding some papers.'; `the cat jumps on the dog.'>}.  
In contrast, we observe that in many cases accuracy and ECE worsen when we filter out possibly mis-labeled samples in the 33\% ambiguous samples by using AUM, suggesting that all ambiguous samples are useful for learning and generalization (which is consistent with \citet{swayamdipta2020dataset}).
Second, we observe that fine-tuning BERT on both the easy-to-learn and the ambiguous samples (\textit{66\% train, Easy-to-learn \& Ambiguous}) achieves similar performance and ECE as \textit{100\% train}.

\paragraph{Main Results} Table \ref{tb:2} shows the result of the comparison of our proposed TDMixUp and baseline methods. 
We observe that our proposed method generally achieves higher accuracy and lower ECE on both in-domain and out-of-domain settings compared to any baseline using the full 100\% training data showing the effectiveness of our TDMixUp strategy. 

\subsection{Ablation Study}
To compare the impact of the MixUp operation on samples generated by \textit{random pairing} and on samples generated by \textit{informative pairing}, we conduct an ablation study. Specifically, we compare the results of MixUp on 66\% train set (i.e., conduct the MixUp operation between randomly selected samples on 66\% train set, which is the union of the top 33\% easy-to-learn and the top 33\% ambiguous samples) and our proposed TDMixUp (i.e., conduct the MixUp operation between the easy-to-learn filtered by AUM and the ambiguous samples). As shown in Table \ref{tb:3}, we observe that our proposed TDMixUp which selects informative samples to combine performs better with respect to accuracy and ECE than vanilla MixUp that selects random samples, in all cases (in-domain and out-of-domain).

\begin{table}[hbt!]
\small
\centering
\begin{tabular}{lcccccc}
\toprule
                              & \multicolumn{1}{c}{Acc} & \multicolumn{1}{c}{ECE} & \multicolumn{1}{c}{Acc} & \multicolumn{1}{c}{ECE} & \multicolumn{1}{c}{Acc} & \multicolumn{1}{c}{ECE} \\ \midrule
                              & \multicolumn{2}{c}{SNLI}                               & \multicolumn{2}{c}{QQP}                                & \multicolumn{2}{c}{SWAG}                               \\ 
\midrule
{Random} & 89.59              & 1.70           & 89.87               & 3.06           &79.15                &4.51  \\
{Ours} & {90.31}               & {1.22}           & {90.42}               & {1.53}           & {79.59}               & {2.16} \\\midrule \midrule
          & \multicolumn{2}{c}{MNLI}                               & \multicolumn{2}{c}{TwitterPPDB}                        & \multicolumn{2}{c}{HellaSWAG}                          \\ \midrule
{Random} & 73.22               & 6.89           & 87.23               & 6.53           &34.43                &15.87        \\
{Ours} & {74.28}               & {2.91}           & {87.89}               & {6.08}           & {35.21}               & {9.45}           \\ \bottomrule
\end{tabular}
\caption{\small{The results comparison of MixUp selecting random samples on the union of the top 33\% easy-to-learn and the top 33\% ambiguous samples (i.e., Random) and our method.}}
\label{tb:3}
\end{table}

\section{Conclusion}
In this work, we propose a novel MixUp that leverages training dynamics (confidence, variability, and Area Under the Margin) to allow more informative samples to be combined in generating augmented samples. We empirically validate that our method not only achieves competitive accuracy but also calibrates BERT model on various NLP tasks, both on in-domain and out-of-domain settings.

\section*{Acknowledgements}
This research is supported in part by NSF CAREER award \#1802358 and NSF CRI award \#1823292. Any opinions, findings, and conclusions expressed here are those of the authors and do not necessarily reflect the views of NSF. We thank AWS for computing resources. We also thank our anonymous reviewers for their constructive feedback. % and comments, which helped improve our paper. 

\newpage
\bibliography{custom}
\bibliographystyle{acl_natbib}

%\newpage
\appendix
\section{Supplementary Materials}
\label{sec:appendix}

\begin{figure}[t]
    \centering
    \includegraphics[scale=0.38]{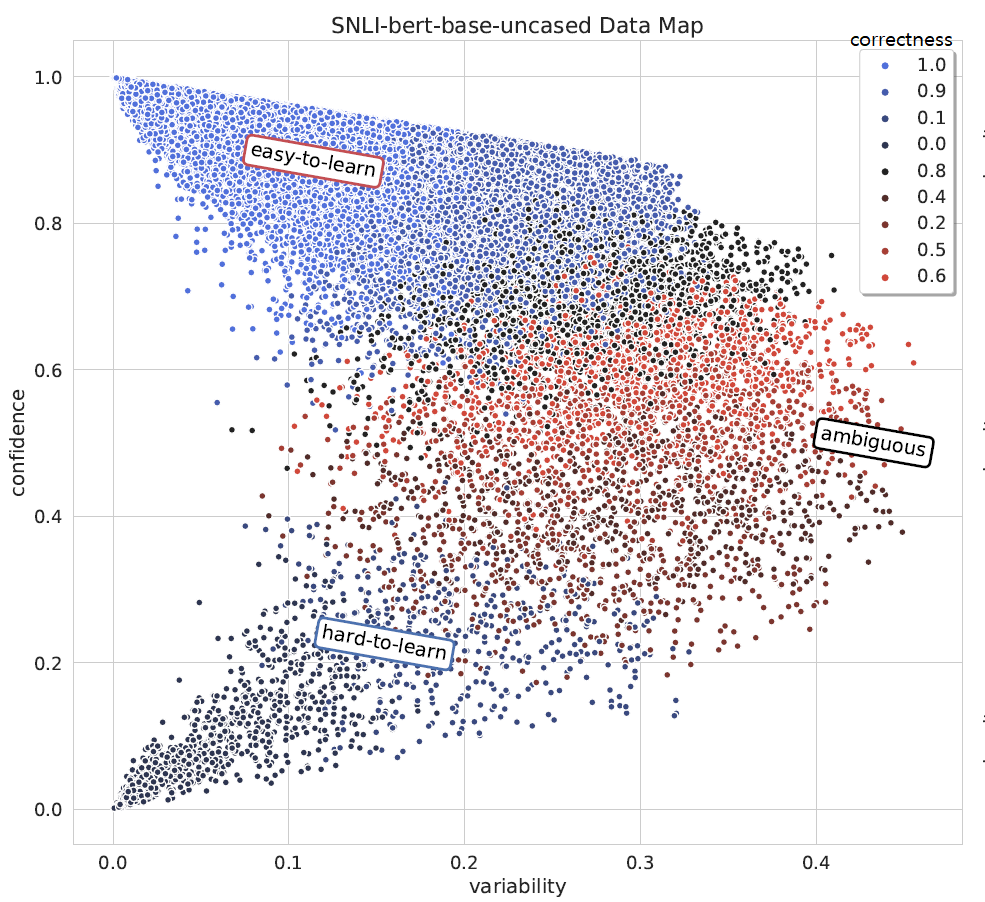}
    \includegraphics[scale=0.38]{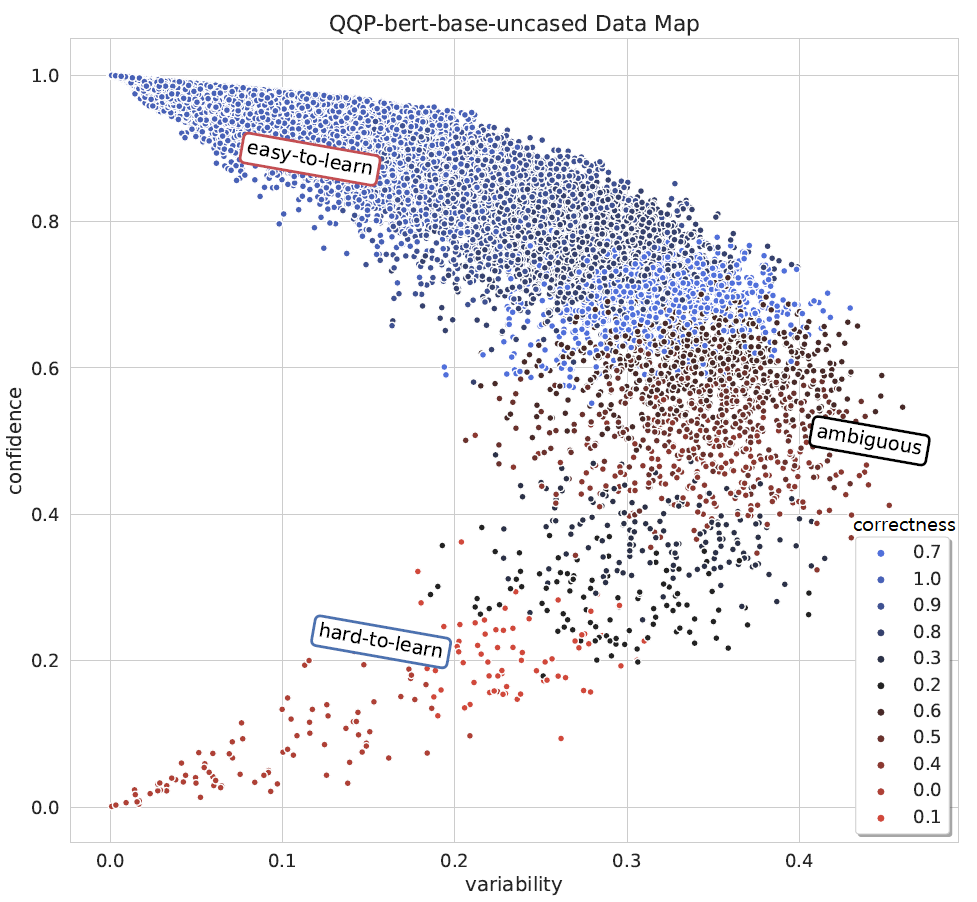}
    \includegraphics[scale=0.38]{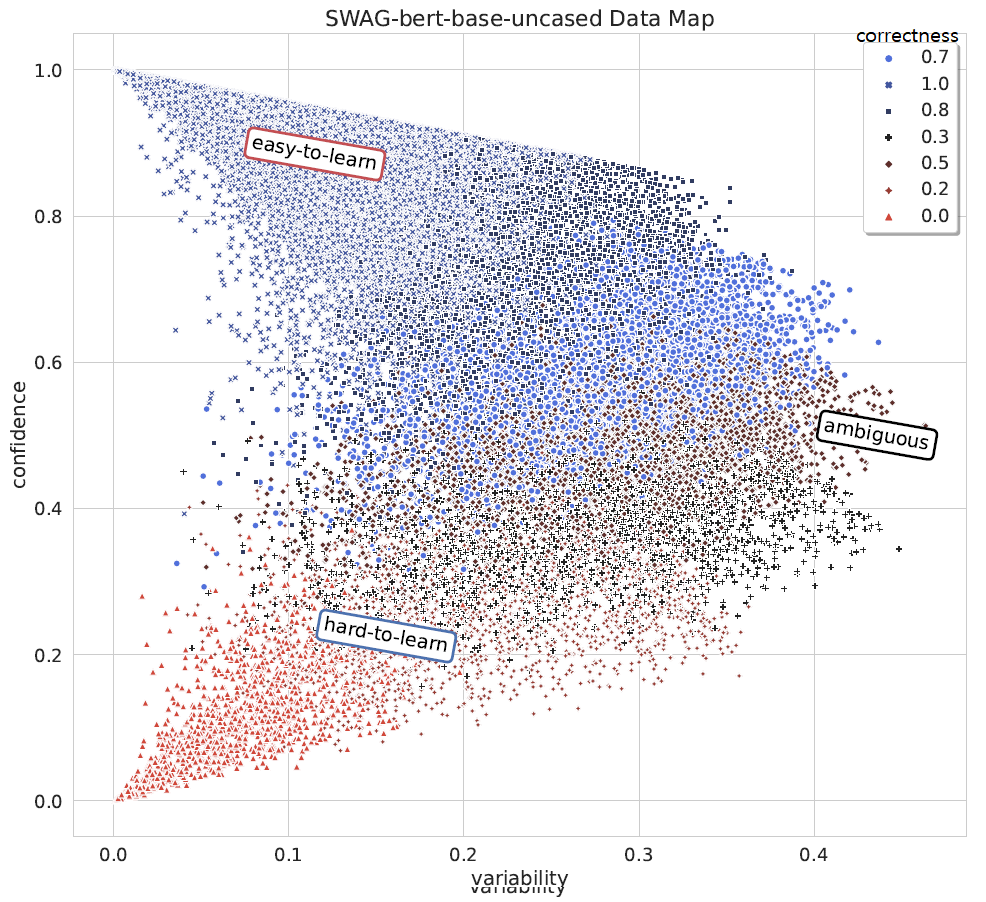}
    \caption{Data Maps of SNLI, QQP and SWAG on \textit{bert-bert-uncased} model.}
    \label{fig}
\end{figure}
\subsection{Training Details}
In our experiments, we use \textit{bert-base-uncased} classification model on top of a task-specific fully-connected layer. The model is fine-tuned with a maximum of 3 epochs, batch size of 16 for SNLI and QQP, batch size 4 for SWAG, a learning rate of 1e-5, gradient clip of 1.0, and no weight decay. We use the hyper-parameter of MixUp $\alpha$ as 0.4. All hyper-parameters are estimated on the validation set of each task. For all results, we report averaged results across 5 fine-tuning runs with random starts. Finally, all experiments are conducted on a single NVIDIA RTX A6000 48G GPU with the total time for fine-tuning all models being under 24 hours. 
For each dataset, we follow the published train/validation/test split by \citet{desai2020calibration} and show the statistics of the datasets in Table \ref{tb:4}.
\begin{table}[h]
    \centering
    \begin{tabular}{lccc}
    \toprule
        Dataset&	Train&	Dev&	Test \\ \midrule
        SNLI&	549,368&	4,922&	4,923 \\
        MNLI&	392,702&	4,908&	4,907 \\
        QQP&	363,871&	20,216&	20,217 \\
        TwitterPPDB&	46,667&	5,060&	5,060 \\
        SWAG&	73,547&	10,004&	10,004 \\
        HellaSWAG&	39,905&	5,021&	5,021 \\
    \bottomrule
    \end{tabular}
    \caption{The statistics of all used datasets.}
    \label{tb:4}
\end{table}
\subsection{Data Maps}
In this section, we provide data maps \cite{swayamdipta2020dataset} of our in-domain datasets on \textit{bert-base-uncased} model in Figure \ref{fig}. These data maps are used to identify the characteristics of each training sample (i.e., easy-to-learn, ambiguous, and hard-to-learn).

\end{document}